\documentclass[10pt,twocolumn,letterpaper]{article}

 \usepackage[pagenumbers]{cvpr} 

\usepackage{graphicx}
\usepackage[dvipsnames]{xcolor}
\usepackage{amsmath}
\usepackage{amssymb}
\usepackage{booktabs}
\usepackage{pifont}
\newcommand{\xmark}{\ding{55}}%
\newcommand{\cmark}{\ding{51}}%
\usepackage{bm}
\newcommand{\bx}{{\bm x}}
\usepackage[accsupp]{axessibility}
\usepackage{float}
\usepackage[export]{adjustbox}
\usepackage{graphics}
\usepackage{makecell}
\usepackage{multirow}
\usepackage{capt-of,etoolbox}

\usepackage[pagebackref,breaklinks,colorlinks]{hyperref}

\usepackage[capitalize]{cleveref}
\crefname{section}{Sec.}{Secs.}
\Crefname{section}{Section}{Sections}
\Crefname{table}{Table}{Tables}
\crefname{table}{Tab.}{Tabs.}

\def\cvprPaperID{10} 
\def\confName{CVPRW}
\def\confYear{2023}

\begin{document}



\title{SPARTAN: Self-supervised Spatiotemporal Transformers Approach to \\ Group Activity Recognition\vspace{-6mm}}

\author{Naga VS Raviteja Chappa$^{1}$, Pha Nguyen$^{1}$, Alexander H Nelson$^{1}$, Han-Seok Seo$^{1}$, Xin Li$^{5}$ \\Page Daniel Dobbs$^{1}$,Khoa Luu$^{1}$\\
$^{1}$University of Arkansas
$^{5}$West Virginia University\\
\tt\small \{nchappa, panguyen, ahnelson, hanseok, pdobbs, khoaluu\}@uark.edu,  
\tt\small Xin.Li@mail.wvu.edu \\
\url{https://uark-cviu.github.io}
\vspace{-2mm}
}
\maketitle

\begin{abstract}

In this paper, we propose a new, simple, and effective Self-supervised Spatio-temporal Transformers (SPARTAN) approach to Group Activity Recognition (GAR) using unlabeled video data. Given a video, we create local and global Spatio-temporal views with varying spatial patch sizes and frame rates. The proposed self-supervised objective aims to match the features of these contrasting views representing the same video to be consistent with the variations in spatiotemporal domains. To the best of our knowledge, the proposed mechanism is one of the first works to alleviate the weakly supervised setting of GAR using the encoders in video transformers. Furthermore, using the advantage of transformer models, our proposed approach supports long-term relationship modeling along spatio-temporal dimensions. The proposed SPARTAN approach performs well on two group activity recognition benchmarks, including NBA and Volleyball datasets, by surpassing the state-of-the-art results by a significant margin in terms of MCA and MPCA metrics\footnote{The implementation of SPARTAN is available at \url{https://github.com/uark-cviu/SPARTAN}}.

\end{abstract}
\vspace{-2mm}
\section{Introduction}
\label{sec:intro}
Group Activity Recognition (GAR) aims to classify the collective actions of individuals in a video clip. This field has gained significant attention due to its diverse applications such as sports video analysis, video monitoring, and interpretation of social situations. 
Diverging significantly from the traditional methods of action recognition that concentrate on comprehending indivudal actions~\cite{wang2016temporal, carreira2017quo, wang2018non, Ranasinghe_2022_CVPR}, whereas the GAR requires a fine-grained analysis of multiple actors interactions in a given scene. Thereby providing a challenging scenario which include the consistency of interaction analysis in spatial and temporal domains and precise identification of the corresponding actors.
Considering the discussed challenges, the necessity of actors' ground-truth bounding boxes is inherent during training and testing and their corresponding action labels during only training phase for most of the current GAR methods~\cite{ibrahim2016hierarchical, wu2019learning, hu2020progressive, gavrilyuk2020actor, pramono2020empowering, ehsanpour2020joint, yan2020higcin, yuan2021learning, li2021groupformer, quach2022non, 9897440}. By utilizing approaches like \emph{RoIPool}~\cite{ren2015faster} and \emph{RoIAlign}~\cite{he2017mask}, the actor features are extracted using the bounding box information which aids to the accurate understanding of their spatiotemporal relationships. To perform the group activity classification, video representations are formed at the group level by combining all the extracted individual actor features while considering the inter-actor relationships, which are employed as input to a classifier.

\begin{figure}[!t]
    \centering
\includegraphics[width=0.46\textwidth]{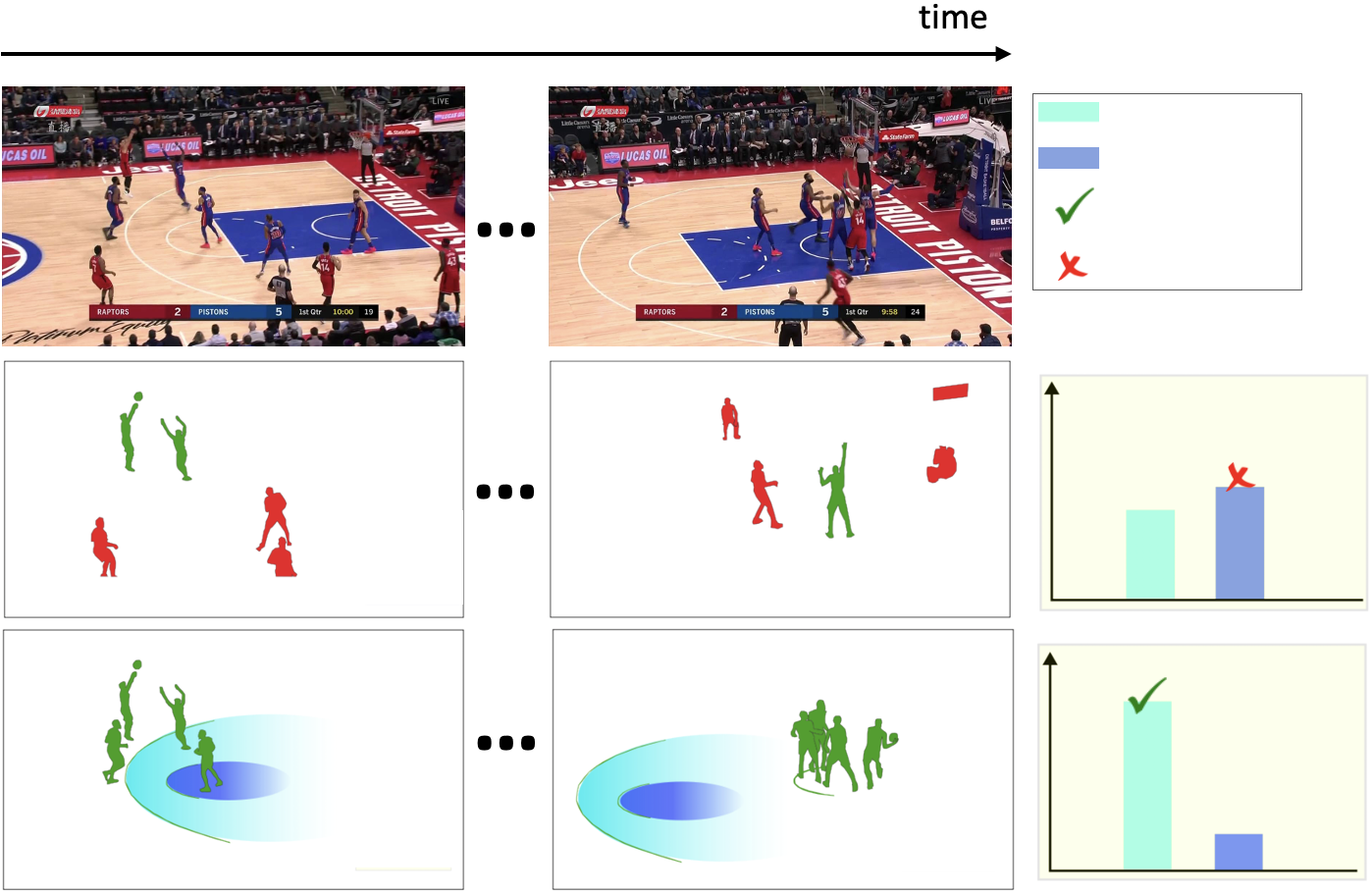}
\put(-55,95){\tiny Predicted confidence scores} 
\put(-41,90){\tiny per category}
\put(-46,103){\tiny False prediction}
\put(-46,113){\tiny True prediction}
\put(-45,121){\tiny 2p.-succ.}
\put(-45,128){\tiny 3p.-succ.}
\put(-27,82){\tiny DFWSGAR}
\put(-24,37){\tiny SPARTAN}
\put(-150, -8){(i)}
\put(-35, -8){(ii)}

    \caption{Visualization of attention captured by the model. (i) The attention in this example focuses on how the relationship is established between the actors. Original sequence from NBA dataset~\cite{yan2020social} (top), Attention captured by DFWSGAR~\cite{kim2022detector} (middle), and SPARTAN model (bottom). \textcolor{red}{Red}-colored actors are the irrelevant information to determine the group activity, whereas \textcolor{ForestGreen}{green}-colored actors, including their positions, are the most relevant. (ii) illustrates that DFWSGAR predicts the category wrong due to the effects shown in (i) whereas SPARTAN is more confident in the prediction, which is further justified by the t-SNE plot as shown in \cref{fig:tsne}.}
     \vspace{-0.2in}
    \label{fig:page1}
\end{figure}
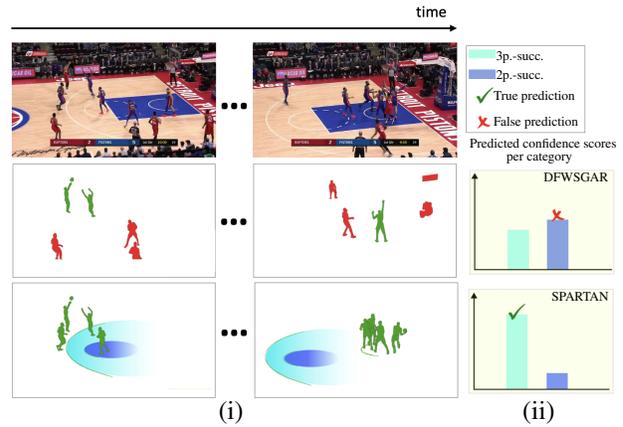
Despite the fact that these approaches performed admirably on the difficult task, their reliance on bounding boxes at inference and substantial data labelling annotations makes them unworkable and severely limits their application.
%
A potential solution to address this issue is simultaneous training of detecting the actors and performing the classification of group activity utilizing all the action labels and bounding box information~\cite{bagautdinov2017social, zhang2019fast}. However, this method implemented training in a fully-supervised setting. Yan~\etal~\cite{yan2020social} presented the Weakly Supervised GAR (WSGAR) learning approach, in the process of mitigating the usage of fine-grained labels during both upstream and downstream tasks.
They generate actor box suggestions using a detector that has been pre-trained on an external dataset in order to solve the absence of bounding box labels. They then learn to eliminate irrelevant possibilities.
Recently, Kim~\etal~\cite{kim2022detector} introduced a detector-free method for WSGAR task which captures the actor information using partial contexts of the token embeddings.
However, the previous methods~\cite{yan2020social, kim2022detector} have various drawbacks as follows. 
First, a detector~\cite{yan2020social} often leads to missing detection of people in case of occlusion, which minimizes overall accuracy. 
Second, partial contexts~\cite{kim2022detector} can only learn if and only if there is movement in consecutive frames. This can be inferred from the illustration in \cref{fig:page1}. 
Third, the temporal information among the tokens must be consistent, and \cite{kim2022detector} does not consider different tokens.

In this paper, we introduce a new simple but effective Self-Supervised \textbf{Spa}tio-tempo\textbf{r}al \textbf{T}r\textbf{an}sformers (SPARTAN) approach to the task of Group Action Recognition that is independent of ground-truth bounding boxes, labels during pre-training, and object detector. Our mechanism only exploits the motion as a supervisory signal from the RGB data modality. 
As seen in Fig.~\ref{fig:page1} (i), our model captures not only the key actors but also their positions, which shows that our method is more effective in group activity classification than DFWSGAR~\cite{kim2022detector}. Our approach is designed to benefit from varying spatial and temporal details within the same deep network. We use a video transformer \cite{gberta_2021_ICML} based approach to handle varying temporal resolutions within the same architecture. Furthermore, the self-attention mechanism in video transformers can capture local and global long-range dependencies in both space and time, offering much larger receptive fields compared to standard convolutional kernels \cite{naseer2021intriguing}. 
%
%
%
The contributions of this work can be summarized as follows.
\begin{itemize}
    \item Instead of considering only motion features across consecutive frames \cite{kim2022detector}, we introduce the first training approach to GAR by exploiting spatial-temporal correspondences. The proposed method varies the space-time features of the inputs to learn long-range dependencies in spatial and temporal domains. 

    \item A new self-supervised learning strategy is performed by jointly learning the inter-frame, i.e., frame-level \textit{temporal}, and intra-frame, i.e., patch-level \textit{spatial}, correspondences further forming into \emph{Inter Teacher-Inter Student loss} and \emph{Inter Teacher-Intra Student loss}. In particular, the spatiotemporal features global, from the entire sequence, and local from the sampled sequence are matched by the learning objectives of the frame level and the patch level in the latent space.
    
    
    \item With extensive experiments on NBA\cite{yan2020social}, and Volleyball\cite{ibrahim2016hierarchical} datasets, the proposed method shows the State-of-the-Art (SOTA) performance results using only RGB inputs.
\end{itemize}

\begin{figure*}
    \centering
    \includegraphics[width=0.90\textwidth]{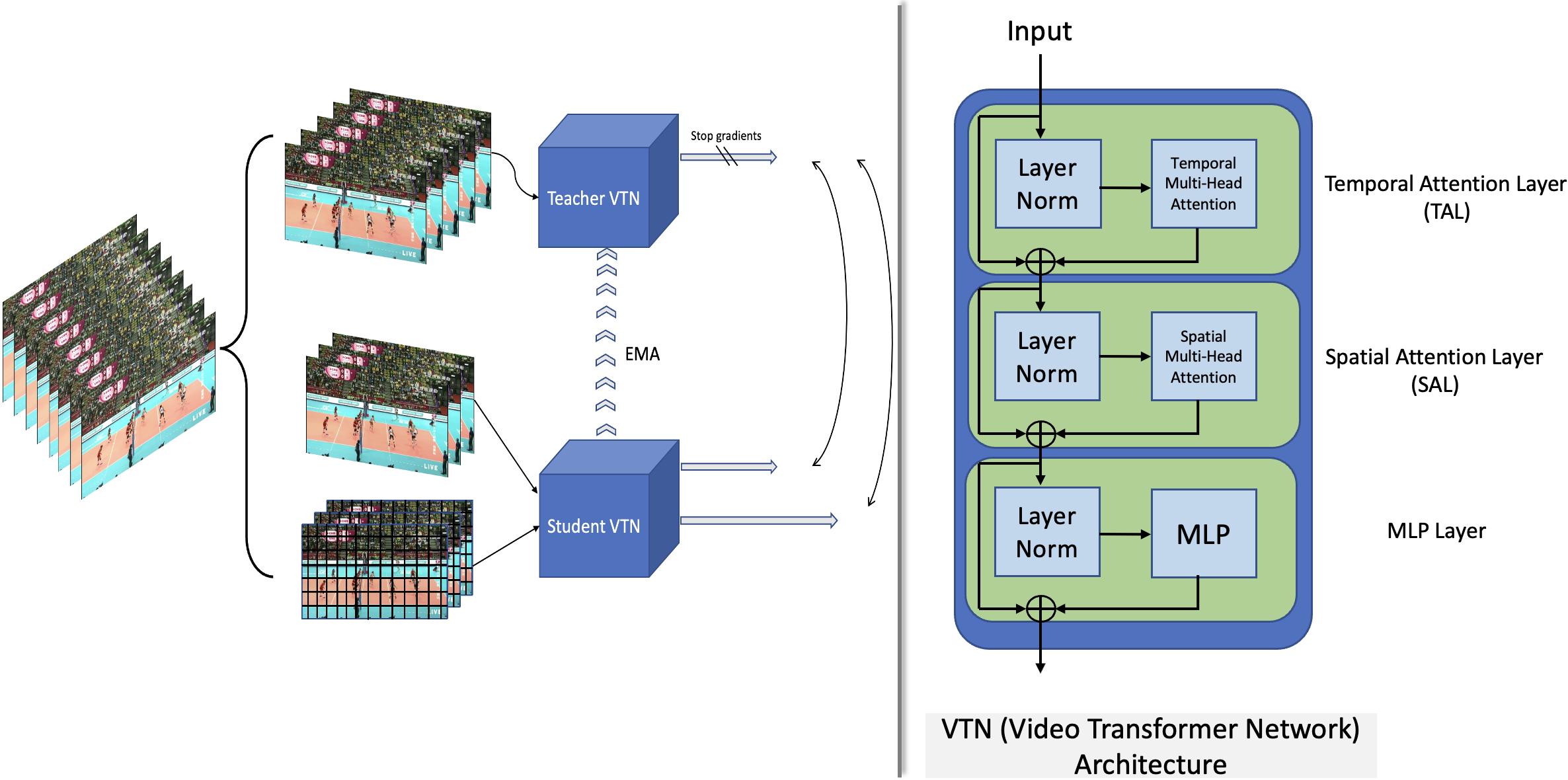}
    \put(-330,18){$\mathcal{L}_{g_{t}-l_{t}}$:\scriptsize Inter Teacher-Inter Student Loss}
    \put(-330,04){$\mathcal{L}_{g_{t}-l_{s}}$:\scriptsize Inter Teacher-Intra Student Loss}
    \put(-235, 145){\rotatebox[origin=c]{90}{$\mathcal{L}_{g_{t}-l_{t}}$}}
    \put(-221, 145){\rotatebox[origin=c]{90}{$\mathcal{L}_{g_{t}-l_{s}}$}}
    \put(-245,190){\scriptsize $\bm{\Tilde{f}_{g_{t}}}$}
    \put(-245,93){\scriptsize $\bm{\Tilde{f}_{l_{t}}}$}
    \put(-225,75){\scriptsize $\bm{\Tilde{f}_{l_{s}}}$}
    \put(-167,26){$\scriptsize \bm{\Tilde{f}_{x}}$}
    \put(-408,181){$\bm{g}_{t}$}
    \put(-404,110){$\bm{l}_{t}$}
    \put(-404,68){$\bm{l}_{s}$}
    \put(-430, 235){\scriptsize Global}
    \put(-440, 225){\scriptsize Temporal Views}
    \put(-452, 215){\scriptsize \textbf{[B$\times$$K_{g}$$\times$C$\times$$H_{g}$$\times$$W_{g}$]}}
    \put(-410, 40){\scriptsize Local}
    \put(-430, 31){\scriptsize Spatiotemporal Views}
    \put(-430, 20){\scriptsize \textbf{[B$\times$$K_{l}$$\times$C$\times$$H_{l}$$\times$$W_{l}$]}}
    \put(-485, 78){\scriptsize Input Video}
    \put(-450,78){\scriptsize ($\bm{X}$)}
    \put(-488, 70){\scriptsize \textbf{[B$\times$T$\times$C$\times$H$\times$W]}}
    \caption{\textbf{The proposed SPARTAN Framework} samples gave input video into global and local views. The sampling strategy for video clips results in different frame rates and spatial characteristics between global views and local views, which are subject to spatial augmentations and have limited fields of view. The teacher model processes global views ($\bm{g}_{t}$) to generate a target, while the student model processes local views ($\bm{l}_{t}$ \& $\bm{l}_{s}$) where $K{l}$ $\le$ $K_{g}$. The network weights are updated by matching the online student local views to the target teacher global views, which involves \emph{cross-view correspondences} and \emph{motion correspondences}. Our approach utilizes a standard ViT backbone with separate space-time attention \cite{gberta_2021_ICML} and an MLP to predict target features from online features.}
    \label{fig:framework}
\end{figure*}
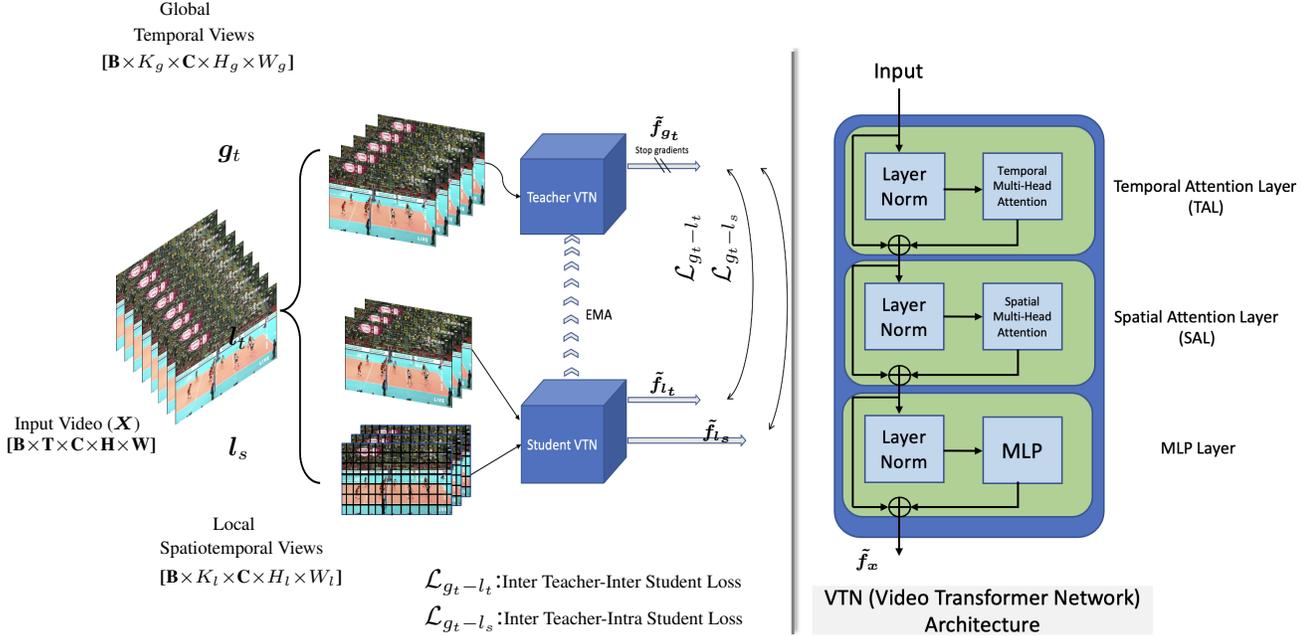
\vspace{-3mm}
\section{Related Work}

\subsection{Group Activity Recognition (GAR)}
Due to the wide range of applications, GAR has recently gained more attention. The initial approaches in the field utilized probabilistic graphical methods ~\cite{amer2014hirf,amer2013monte,amer2015sum,lan2012social,lan2011discriminative,wang2013bilinear} and AND-OR grammar methods ~\cite{amer2012cost,shu2015joint} to process the extracted features. 
As deep learning evolved over the years, methods involving Convolutional Neural Networks (CNN) ~\cite{bagautdinov2017social,ibrahim2016hierarchical}, Recurrent Neural Networks (RNN) ~\cite{wang2017recurrent,yan2018participation,qi2018stagnet,bagautdinov2017social,deng2016structure,shu2019hierarchical,li2017sbgar,ibrahim2016hierarchical,ibrahim2018hierarchical} achieved outstanding performance thanks to their learning power of high-level information and temporal context.

Recent methods for identifying group actions~\cite{wu2019learning,gavrilyuk2020actor,hu2020progressive,yan2020social,ehsanpour2020joint,pramono2020empowering,li2021groupformer,yuan2021spatio} typically utilize attention-based models and require explicit character representations to model spatial-temporal relations in group activities. Graph convolution networks, as described in~\cite{wu2019learning,yuan2021spatio}, are used to learn spatial and temporal information of actors by constructing relational graphs, while Rui~\etal~\cite{yan2020social} suggest building spatial and temporal relation graphs to infer actor links. Kirill~\etal~\cite{gavrilyuk2020actor} use a transformer encoder-based technique with different backbone networks to extract features for learning actor interactions from multimodal inputs. Li~\etal~\cite{li2021groupformer} use a clustered attention approach to capture contextual spatial-temporal information.
Mingfei~\etal~\cite{han2022dual} proposed MAC-Loss which is a combination of spatial and temporal transformers in two complimentary orders to enhance the learning effectiveness of actor interactions and preserve actor consistency at the frame and video levels. 
\textbf{Weakly supervised group activity recognition (WSGAR).} Several techniques have been developed to tackle WSGAR with limited supervision, like training detectors within the framework using bounding boxes. One approach is WSGAR, which does not rely on bounding box annotations during training or inference and incorporates an off-the-shelf item detector into the model. Another technique, proposed by Zhang et al. \cite{zhang2021multi}, uses activity-specific characteristics to improve WSGAR, but is not specifically designed for GAR. Kim et al. \cite{kim2022detector} proposed a detector-free method that uses transformer encoders to extract motion features. Our proposed method is a self-supervised training approach dedicated to WSGAR, which does not require actor-level annotations, object detectors, or labels.

\noindent\textbf{Transformers in Vision}. 
Vaswani~\etal~\cite{vaswani2017attention} introduced the transformer architecture for sequence-to-sequence machine translation. This architecture has since been widely adopted to many various natural processing tasks. 
Dosovitskiy~\etal~\cite{dosovitskiy2020image} introduced a transformer architecture that is not based on convolution for image recognition tasks. For different downstream computer vision tasks, these works~\cite{li2021ffa, yuan2021tokens,liu2021swin,wang2021pyramid} used transformer architecture as a general backbone to make exceptional performance progress. In the video domain, many works~\cite{han2020mining, arnab2021vivit, li2022uniformer, bertasius2021space,fan2021multiscale,patrick2021keeping, Quach_2021_CVPR, 9895423, quach2022depth} exploited spatial and temporal self-attention to learn video representation efficiently. 
Patrick~\etal~\cite{patrick2021keeping} introduce a self-attention block that focuses on the trajectory, which tracks the patches of space and time in a video transformer.

\section{SPARTAN}

The proposed method aims to recognize a group activity in a given video in the absence of the ground-truth information like actor corresponding bounding boxes. The general architecture of our self-supervised training within the teacher-student framework for group activity recognition is illustrated in Fig. \ref{fig:framework}. Unlike the other contrastive learning methods, we process two clips from the same video by changing their spatial-temporal characteristics, which do not rely on the memory banks. The proposed loss formulation matches the features of the two dissimilar clips to impose consistency in motion and spatial changes in the same video. The proposed SPARTAN framework will be discussed further in the following sections.
\subsection{Self-Supervised Training}\label{subsec:ssltraining}
Given the high temporal dimensionality of videos, motion and spatial characteristics of the group activity will be learned, such as 3p.-succ. (from NBA dataset~\cite{yan2020social}) or l-spike (from Volleyball dataset~\cite{ibrahim2016hierarchical}) during the video. Thus, several video clips with different motion characteristics can be sampled from a single video. A key novelty of the proposed approach involves predicting these different video clips with varying temporal characteristics from each other in the feature space. It leads to learning contextual information that defines the underlying distribution of videos and makes the network invariant to motion, scale, and viewpoint variations. Thus, self-supervision for video representation learning is formulated as a motion prediction problem that has three key components: \textbf{a)} We generate multiple temporal views consisting of different numbers of clips with varying motion characteristics from the same video as in \cref{subsec:mo_pred}, \textbf{b)} In addition to motion, we vary spatial characteristics of these views as well by generating local, i.e., smaller spatial field, and global, i.e., higher spatial field, of the sampled clips as in \cref{subsec:cross_view_corrspondences}, and \textbf{c)} We introduce a loss function in \cref{subsec:loss} that matches the varying views across spatial and temporal dimensions in the latent space.
\vspace{-3mm}
\subsubsection{Motion Prediction as Self-supervision Learning}
\label{subsec:mo_pred}
The frame rate is a crucial aspect of a video as it can significantly alter the motion context of the content. For instance, the frame rate can affect the perception of actions, such as walking slowly versus walking quickly, and can capture subtle nuances, such as the slight body movements in walking. Traditionally, video clips are sampled at a fixed frame rate \cite{qian2020spatiotemporal, xiao2021modist}. However, when comparing views with different frame rates, i.e., varying numbers of clips, predicting one view from another in feature space requires explicitly modeling object motion across clips. Furthermore, predicting subtle movements captured at high frame rates compels the model to learn contextual information about motion from a low frame rate input.
 
\textbf{Temporal Views:} 
We refer to a collection of clips sampled at a specific video frame rate as a temporal view. We generate different views by sampling at different frame rates, producing temporal views with varying resolutions. The number of temporal tokens ($T$) input to ViT varies in different views. Our proposed method enforces the correspondences between such views, which allows for capturing different motion characteristics of the same action. We randomly sampled these views to create motion differences among them. Our ViT models process these views, and we predict one view from the other in the latent space. In addition to varying temporal resolution, we also vary the resolution of clips across the spatial dimension within these views. It means that the spatial size of a clip can be lower than the maximum spatial size (224), which can also decrease the number of spatial tokens.   Similar sampling strategies have been used \cite{feichtenhofer2019slowfast, kahatapitiya2021coarse} but under multi-network settings, while our approach handles such variability in temporal resolutions with a single ViT model by using vanilla positional encoding~\cite{vaswani2017attention}. 
\vspace{-2mm}
\subsubsection{Cross-View Correspondences}
\label{subsec:cross_view_corrspondences}
Our training strategy aims to learn the relationships between a given video's temporal and spatial dimensions. To this end, we propose novel cross-view correspondences by altering the field of view during sampling. We generated global and local temporal views from a given video to achieve this.

\textbf{Global Temporal Views ($\bm{g}_{t}$):} We randomly sample $K_{g}$ (is equal to $T$) frames from a video clip with spatial size fixed to $W_{global}$ and $H_{global}$. These views are fed into the teacher network which yields an output denoted by $\bm{\Tilde{f}_{g_{t}}}$.

\textbf{Local Spatiotemporal Views ($\bm{l}_{t}$ and $\bm{l}_{s}$):} 
Local views cover a limited portion of the video along both spatial and temporal dimensions. We generate local temporal views by randomly sampling several frames $K_{l}$ ($\leq$ $K_{g}$) with a spatial size fixed to $W_{local}$ and $H_{local}$. These views are fed into the student network which yields two outputs denoted by $\bm{\Tilde{f}_{l_{t}}}$ and $\bm{\Tilde{f}_{l_{s}}}$ respectively.

\textbf{Augmentations:} 
We apply different data augmentation techniques to the spatial dimension, that is, to the clips sampled for each view. Specifically, we apply color jittering and gray scaling with probability 0.8 and 0.2, respectively, to all temporal views. We apply Gaussian blur and solarization with probability 0.1 and 0.2, respectively, to global temporal views.

Our approach is based on the intuition that learning to predict a global temporal view of a video from a local temporal view in the latent space can help the model capture high-level contextual information. Specifically, our method encourages the model to model both spatial and temporal context, where the spatial context refers to the possibilities surrounding a given spatial crop and the temporal context refers to possible previous or future clips from a given temporal crop. It is important to note that spatial correspondences also involve a temporal component, as our approach attempts to predict a global view at timestamp $t=j$ from a local view at timestamp $t=i$. To enforce these cross-view correspondences, we use a similarity objective that predicts different views from each other.

\subsection{The Proposed Objective Function}\label{subsec:loss}
Our model is trained with an objective function that predicts different views from each other. These views represent different spatial-temporal variations that belong to the same video. 

Given a video $\bm{X}=\{\bx_t\}_{t=1}^T$, where $T$ represents the number of frames, let $\bm{g}_{t}$, $\bm{l}_{t}$ and $\bm{l}_{s}$ represent global temporal views, local temporal and spatial views such that $\bm{g}_{t}=\{\bx_t\}_{t=1}^{K_{g}}$ and $\bm{l}_{t} = \bm{l}_{s} =\{\bx_t\}_{t=1}^{K_{l}}$, where $\bm{g}_{t}$, $\bm{l}_{t}$ and $\bm{l}_{s}$ are subsets of video $\bm{X}$ and $K_{l} \le K_{g}$ where $K_{g}$ and $K_{l}$ are the number of frames for teacher and student (global and local) inputs. We randomly sample $K_{g}$ global and $K_{l}$ local temporal views as in \cref{subsec:cross_view_corrspondences}. These temporal views are passed through the student and teacher models to get the corresponding class tokens or feature $\bm{f}_g$ and $\bm{f}_l$. These class tokens are normalized as follows.
\begin{equation}
     \bm{\Tilde{f}}^{(i)} = \frac{\text{exp}(\bm{f}^{(i)}) / \tau}{\sum_{i=1}^n \text{exp}(\bm{f}^{(i)})/ \tau },
\end{equation}
\begin{equation}
\bm{\Tilde{f}}^{(i)} = \frac{\sum_{j=1}^{m}\text{exp}(\beta_j \cdot \bm{f}^{(i)}) / \tau^{\frac{1}{\alpha_j}}}{\sum_{i=1}^{n} \sum_{j=1}^{m} \text{exp}(\beta_j \cdot \bm{f}^{(i)})/ \tau^{\frac{1}{\alpha_j}}},
\end{equation}
where $\tau$ is a temperature parameter used to control the sharpness of the exponential function \cite{caron2021emerging} and $\bm{f}^{(i)}$ is each element in $\bm{\Tilde{f}^{(i)}}\in\mathbb{R}^{n}$.  

\textbf{Inter Teacher-Inter Student Loss:} 
Our $\bm{g}_{t}$ have the same spatial size but differ in temporal content because the number of clips/frames is randomly sampled for each view. One of the $\bm{g}_{t}$ always passes through the teacher model that serves as the target label. We map the student's $\bm{l}_{t}$ with the teacher's $\bm{g}_{t}$ to create a global-to-local temporal loss as in Eqn. \eqref{eq:global_to_global_loss}.
\begin{align}\label{eq:global_to_global_loss}
    \mathcal{L}_{g_{t}-l_{t}} &=  -\bm{\Tilde{f}}_{g_{t}} * log(\bm{\Tilde{f}}_{l_{t}}),
\end{align}
where $\bm{\Tilde{f}_{g_{t}}}$ and $\bm{\Tilde{f}_{l_{t}}}$ are the tokens of the class for $\bm{g}_{t}$ and $\bm{l}_{t}$ produced by the teacher and student, respectively.


\textbf{Inter Teacher-Intra Student Loss:} 
Our $\bm{l}_{t}$ have a limited field of vision along the spatial and temporal dimensions compared to the $\bm{g}_{t}$. However, the number of local views is four times higher than that of global views. All $\bm{l}_{s}$ are passed through the student model and mapped to $\bm{g}_{t}$ from the teacher model to create the loss function as in \cref{eq:local_to_global_loss}.
\begin{align}\label{eq:local_to_global_loss}
    \mathcal{L}_{g_{t}-l_{s}} &= \sum_{n=1}^{q} -\bm{\Tilde{f}}_{g_{t}} * log(\bm{\Tilde{f}}^{(n)}_{l_{s}}),
\end{align}
where $\bm{\Tilde{f}_{l_{s}}}$ are the tokens of the class for $\bm{l}_{s}$ produced by the student and $q$ represents the number of local temporal views set to sixteen in all our experiments. The overall loss to train our model is simply a linear combination of both losses, as in Eqn. \eqref{eq:global_to_global_loss} and Eqn. \eqref{eq:local_to_global_loss}, given as in Eqn. \eqref{eq:totalloss}.
\begin{equation} \label{eq:totalloss}
    \mathcal{L} =  \mathcal{L}_{g_{t}-l_{t}} +  \mathcal{L}_{g_{t}-l_{s}}
\end{equation}
\subsection{Inference}\label{subsec:Inference}

\begin{figure}[!t]
    \centering
    \includegraphics[width=0.47\textwidth]{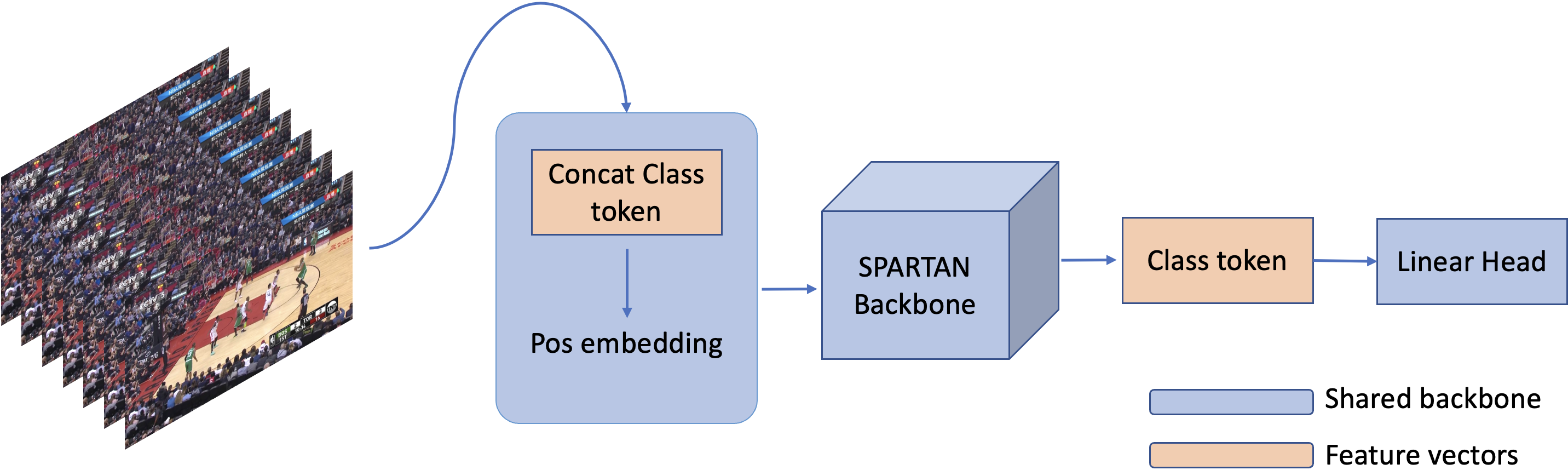}
\caption{\textbf{Inference}. We uniformly sample the video clip and pass it through a shared network and generate feature vectors (class tokens). These vectors are fed to the downstream task classifier.}
    \vspace{-0.1in}
    \label{fig:my_label1}
\end{figure}

Fig. \ref{fig:my_label1} illustrates our inference framework. During this stage, fine-tuning of the trained self-supervised model is performed. We use the pre-trained SPARTAN model and fine-tune the model with the available labels, followed by a linear classifier. We use this on downstream tasks to improve performance.

\section{Experiments}

\subsection{Datasets}
\noindent\textbf{Volleyball Dataset}\cite{ibrahim2016hierarchical} 
comprises 3,493 training and 1,337 testing clips, totaling 4,830 labeled clips, from 55 videos. The dataset contains annotations for eight group activity categories and nine individual action labels with corresponding bounding boxes. However, in our WSGAR experiments, we only use the group activity labels and ignore the individual action annotations. For evaluation, we use Multi-class Classification Accuracy (MCA) and Merged MCA metrics, where the latter merges the right set and right pass classes into right pass-set and the left set and left pass classes into left pass-set, as in previous works such as SAM~\cite{yan2020social} and DFWSGAR~\cite{kim2022detector}. This is done to ensure a fair comparison with existing methods.

\noindent\textbf{NBA Dataset}\cite{yan2020social} 
in our experiment comprises a total of 9,172 labeled clips from 181 NBA videos, with 7,624 clips used for training and 1,548 for testing. Each clip is annotated with one of nine group activities, but there is no information on individual actions or bounding boxes. In evaluating the model, we use the Multi-class Classification Accuracy (MCA) and Mean Per Class Accuracy (MPCA) metrics, with MPCA used to address the issue of class imbalance in the dataset.

\subsection{Deep Network Architecture}
Our video processing approach uses a vision transformer (ViT) \cite{gberta_2021_ICML} to apply individual attention to both the temporal and spatial dimensions of the input video clips. The ViT consists of 12 encoder blocks and can process video clips of size $(B\times T\times C\times W\times H)$, where $B$ and $C$ represent the batch size and the number of color channels, respectively. The maximum spatial and temporal sizes are $W=H=224$ and $T=18$, respectively, meaning that we sample 18 frames from each video and rescale them to $224\times 224$.
Our network architecture (see Fig.~\ref{fig:framework}) is designed to handle variable input resolution during training, such as differences in frame rate, number of frames in a video clip, and spatial size. However, each ViT encoder block processes a maximum of 196 spatial and 16 temporal tokens, and each token has an embedding dimension of $\mathbb{R}^{m}$ \cite{dosovitskiy2020image}. Along with these spatial and temporal input tokens, we also use a single classification token as a characteristic vector within the architecture \cite{devlin2018bert}. This classification token represents the standard features learned by the ViT along the spatial and temporal dimensions of a given video.
During training, we use variable spatial and temporal resolutions that are $W \le 224$, $H \le 224$, and $T \le 18$, which result in various spatial and temporal tokens. Finally, we apply a projection head to the class token of the final ViT encoder \cite{caron2021emerging, grill2020bootstrap}.
 
\textbf{Self-Distillation.} 
In our approach (shown in \cref{fig:framework}), we adopt a teacher-student setup for self-distillation inspired by \cite{caron2021emerging, grill2020bootstrap}. The teacher model has the same architecture as the student model, including the ViT backbone and predictor MLP, but it does not undergo direct training. Instead, during each training step of the student model, we update the teacher weights using an exponential moving average (EMA) of the student weights \cite{caron2021emerging}. This approach enables us to use a single shared network to process multiple input clips.

\subsection{Implementation Details }
For both the NBA and Volleyball datasets, frames are sampled at a rate of T ($K_{g}$) using segment-based sampling \cite{wang2016temporal}. The frames are then resized to $W_{g} = 224$ \& $H_{g} = 224$ for the teacher input and $W_{l} = 96$ \& $H_{l} =96$ for the student input, respectively. For the Volleyball dataset, we use $K_{g}$ = 5 ($K_{l}\in{3,5}$), while for the NBA dataset, we use $K_{g}$ = 18 ($K_{l}\in{2,4,8,16,18}$). We randomly initialize weights relevant to temporal attention, while spatial attention weights are initialized using a ViT model trained self-supervised over ImageNet-1K \cite{imagenet}. This initialization setup allows us to achieve faster convergence of space-time ViT similar to the supervised setting \cite{gberta_2021_ICML}. We use an Adam optimizer \cite{kingma15adam} with a learning rate of $5\times10^{-4}$, scaled using a cosine schedule with a linear warm-up for five epochs \cite{Steiner2021HowTT, chen2021mocov3}. We also use weight decay scaled from 0.04 to 0.1 during training. For the \textbf{downstream task}, we train a linear classifier on our pretrained SPARTAN backbone. During training, the backbone is frozen, and the classifier is trained for 100 epochs with a batch size of 32 on a single NVIDIA-V100 GPU using SGD with an initial learning rate of 1e-3 and a cosine decay schedule. We also set the momentum to 0.9.

 \begin{table}[!t]

\vspace{-4mm}
\begin{center}
\begin{tabular}{>{\arraybackslash}m{2.35cm} | >{\centering\arraybackslash}m{1.3cm}>{\centering\arraybackslash}m{0.85cm}>{\centering\arraybackslash}m{0.8cm}>{\centering\arraybackslash}m{0.9cm}}
\hline
Method                                                        & MCA       & MPCA      \\
\hline
\multicolumn{3}{c}{\textbf{Video backbone}} \\
\hline
TSM~\cite{lin2019tsm}                                        & 66.6      & 60.3      \\
VideoSwin~\cite{liu2021video}                                & 64.3      & 60.6      \\
\hline
\multicolumn{3}{c}{\textbf{GAR model}} \\
\hline
ARG~\cite{wu2019learning}                                      & 59.0      & 56.8      \\
AT~\cite{gavrilyuk2020actor}                                  & 47.1      & 41.5      \\
SACRF~\cite{pramono2020empowering}                            & 56.3      & 52.8      \\
DIN~\cite{yuan2021spatio}                                   & 61.6      & 56.0      \\

SAM~\cite{yan2020social}                          & 54.3      & 51.5      \\

DFWSGAR~\cite{kim2022detector}                                                          & 75.8     & 71.2     \\
\hline
\textbf{Ours}  &\textbf{82.1} & \textbf{72.8}\\
\hline
\end{tabular}
 
\end{center}
\vspace{-4mm}
\caption{Comparisons with the State-of-the-Art GAR models and video backbones on the NBA dataset \cite{yan2020social}.
}
\vspace{-1.5em}

\label{table:SOTA_NBA}
\vspace{-0.5em}
\end{table}

  
\begin{table}[!t]

\vspace{-4mm}
\begin{center}
\begin{tabular}{>{\arraybackslash}m{2.4cm}| >  {\centering\arraybackslash}m{2.3cm} >{\centering\arraybackslash}m{0.9cm}>{\centering\arraybackslash}m{1.0cm}}

\hline
Method                             & Backbone              & MCA   & Merged MCA\\ [0.3ex]
\hline
\multicolumn{4}{c}{\textbf{Fully supervised}} \\
\hline
SSU~\cite{bagautdinov2017social}                 & Inception-v3          & 89.9  & - \\
PCTDM~\cite{yan2018participation}                & ResNet-18             & 90.3  & 94.3\\ %
StagNet~\cite{qi2018stagnet}                     & VGG-16                & 89.3  & - \\
ARG~\cite{wu2019learning}                        & ResNet-18             & 91.1 & \underline{95.1}\\ %
CRM~\cite{azar2019convolutional}        & I3D                   & 92.1  & - \\
HiGCIN~\cite{yan2020higcin}                      & ResNet-18             & 91.4  & - \\
AT~\cite{gavrilyuk2020actor}                     & ResNet-18             & 90.0 & 94.0\\ %
SACRF~\cite{pramono2020empowering}               & ResNet-18             & 90.7 & 92.7   \\ %
DIN~\cite{yuan2021spatio}                        & ResNet-18             & \underline{93.1}  & \textbf{95.6} \\ %
TCE+STBiP~\cite{yuan2021learning}        & VGG-16                & \textbf{94.1}  & - \\
GroupFormer~\cite{li2021groupformer}                      & Inception-v3             & \textbf{94.1}  & - \\

\hline
\multicolumn{4}{c}{\textbf{Weakly supervised}} \\
\hline 
PCTDM~\cite{yan2018participation}               & ResNet-18             & 80.5  & 90.0\\
ARG~\cite{wu2019learning}                       & ResNet-18             & 87.4  & 92.9\\
AT~\cite{gavrilyuk2020actor}                    & ResNet-18             & 84.3  & 89.6\\
SACRF~\cite{pramono2020empowering}              & ResNet-18             & 83.3  & 86.1  \\
DIN~\cite{yuan2021spatio}                       & ResNet-18             & 86.5  & 93.1\\
SAM~\cite{yan2020social}                        & ResNet-18             & 86.3  & 93.1\\

DFWSGAR~\cite{kim2022detector}                                         & ResNet-18             & 90.5  & 94.4\\
\hline
\textbf{Ours} & ViT-Base & \textbf{92.9}& \textbf{95.6} \\
\hline
\end{tabular}

\vspace{-5mm}
\end{center}
\caption{Comparison with the state-of-the-art methods on the Volleyball dataset.~\cite{ibrahim2016hierarchical}}
 \vspace{-1.1em}

\label{table:SOTA_Volleyball}
\end{table}

\begin{table*}[!t]
\begin{minipage}{.55\textwidth}
	\centering \small
    
	\setlength{\tabcolsep}{8pt}
	\scalebox{1.0}[1.0]{
	\begin{tabular}{c|c|c|c|c|c}
		\toprule 
		
		$\bm{l_{t}}\to \bm{g_{t}}$  & $\bm{l_{s}}\to\bm{g_{t}}$  & $\bm{l_{s}}\to\bm{l_{t}}$  & $\bm{g_{t}}\to\bm{l_{t}}$ & NBA     & Volleyball  \\  \midrule
		\cmark   & \xmark   & \xmark   & \xmark  & 61.03   & 62.70 \\ 
		\xmark   & \cmark   & \xmark   & \xmark  & 62.59   & 65.40 \\ 
		\cmark   & \cmark   & \xmark   & \xmark  & \textbf{81.20}   & \textbf{90.80} \\
		\cmark   & \cmark   & \cmark   & \xmark  & 72.11   & 77.62 \\
		\cmark   & \cmark   & \xmark   & \cmark  & 78.17   & 85.88 \\
		\xmark   & \xmark   & \cmark   & \cmark  & 64.36   & 71.87 \\ \bottomrule
	\end{tabular}
    
	}
 \caption{\textbf{View Correspondences (VC).} The most optimal combination for predicting view correspondences involves predicting local-to-global (temporal) and local-to-global (spatial) views, outperforming other combinations.}
    \label{tbl:ablation_correspondences}
\end{minipage}
\hfill
\begin{minipage}{.42\textwidth}
	\centering\small
 \vspace{-1.0em}
 
	\setlength{\tabcolsep}{7pt}
	\scalebox{1.0}[1.0]{
	\begin{tabular}{c|c|c|c}
		\toprule
		 
		Spatial  & Temporal & NBA     & Volleyball \\ \midrule
		\cmark   & \xmark   & 69.38   & 78.59   \\ 
		\xmark   & \cmark   & 72.90   & 81.45   \\ 
		\cmark   & \cmark   & \textbf{81.20}   & \textbf{90.80}   \\ \bottomrule
	\end{tabular}}
\caption{\textbf{Spatial vs Temporal variations.}  The best results are achieved by utilizing cross-view correspondences with varying fields of view along both spatial and temporal dimensions. It is observed that temporal variations between views have a greater impact on performance compared to applying only spatial variation.}
    \label{tbl:ablation_st}
\end{minipage}
\vspace{-0.18in}
\end{table*}

\begin{table}[!t]
\begin{center}

\setlength{\tabcolsep}{5pt}
	\scalebox{1.0}[1.0]{
	\begin{tabular}{c|c|c}
		\toprule
		
		   Method   & NBA     & Volleyball  \\  \midrule
		Ours + TIS \cite{qian2020spatiotemporal} & 78.45   & 88.11 \\ 

		Ours + MC & \textbf{81.20}   & \textbf{90.80} \\ \bottomrule
	\end{tabular}}
 \caption{\textbf{Temporal Sampling Strategy 
    }. We evaluate the effectiveness of our proposed temporal sampling strategy, called "\emph{motion correspondences (MC)}" (\cref{subsec:mo_pred}), by comparing it with an alternate approach, the "temporal interval sampler (TIS)" \cite{qian2020spatiotemporal}, used with CNNs under contrastive settings. 
    }
    \label{tbl:ablation_temporal}

\end{center}

\vspace{-1.0em}
\end{table}

\begin{table}[!t]
\begin{center}

\setlength{\tabcolsep}{5pt}
	\scalebox{0.95}[0.95]{
	\begin{tabular}{c|c|c}
		\toprule
		       Patch size   & NBA     & Volleyball  \\  \midrule
		8 & 78.71   & 87.10 \\
        16 & \textbf{81.20}   & \textbf{90.80} \\
	    32    & 72.56   & 79.21 \\ \bottomrule
	\end{tabular}}
 \caption{\textbf{Spatial Augmentations (SA)}: Applying different patch sizes randomly over the spatial dimensions for different views leads to consistent improvements on both NBA and Volleyball datasets. 
}
    \label{tbl:ablation_aug}

\end{center}
\vspace{-1.0em}

\vspace{-1.0em}
\end{table}

\begin{table}[!t]
\begin{center}

\setlength{\tabcolsep}{5pt}
	\scalebox{1.0}[1.0]{
	\begin{tabular}{c|c|c}
		\toprule
		
		Multi-view      & NBA    & Volleyball  \\  \midrule
		\xmark      & 76.17   & 88.35 \\
		\cmark      & \textbf{81.20}   & \textbf{90.80} \\ \bottomrule
	\end{tabular}}
 \caption{\textbf{Inference}: Providing multiple views of different spatiotemporal resolutions to a shared network (multiview) leads to noticeable performance improvements compared to using a single view for both the NBA and Volleyball datasets.}
    \label{tbl:ablation_inf}

\end{center}
\vspace{-2.1em}

\end{table}

\subsection{Comparison with state-of-the-art methods}

\noindent\textbf{NBA dataset}
We compare our approach to the state-of-the-art in GAR and WSGAR, which leverage bounding box recommendations produced by SAM~\cite{yan2020social}, as well as to current video backbones in the weakly supervised learning environment, using the NBA dataset.
We exclusively utilise RGB frames as input for each approach, including the video backbones, to ensure a fair comparison.
Table~\ref{table:SOTA_NBA} lists the findings.
Please take note that the results of SAM~\cite{yan2020social} has been listed from~\cite{kim2022detector}.
With 6.3\%p of MCA and 1.6\%p of MPCA, the proposed method outperforms existing GAR and WSGAR methods by a significant margin.
Additionally, our approach is contrasted with two current video backbones utilised in traditional action detection, ResNet-18 TSM~\cite{lin2019tsm} and VideoSwin-T~\cite{liu2021video}. These strong backbones perform admirably in WSGAR, but ours is the finest.

\noindent
\textbf{Volleyball dataset.}
For the volleyball dataset, we compare our approach to the most recent GAR and WSGAR approaches in two different supervision levels: fully supervised and weakly supervised.
The usage of actor-level labels, such as individual action class labels and ground-truth bounding boxes, in training and inference differs across the two settings.
To have a fair comparison, all the results reported are using only RGB input and ResNet-18 backbone.
Please consider the note that first is from the original papers, and the second is the MCA values of~\cite{yuan2021spatio}.
We eliminate the individual action classification head and substitute an object detector trained on an external dataset for the ground-truth bounding boxes in the weakly supervised situation.
Table~\ref{table:SOTA_Volleyball} presents the results.
Results from earlier techniques in fully supervised and weakly supervised environments are displayed in the first and second sections, respectively.
In weakly supervised conditions, our technique significantly outperforms all GAR and WSGAR models, outperforming them by 2.4\% of MCA and 1.2\% of Merged MCA when compared to the models' utilising ViT-Base backbone.
Our technique outperforms current GAR methods, such as~\cite{bagautdinov2017social, yan2018participation, qi2018stagnet, gavrilyuk2020actor, pramono2020empowering}, by employing more thorough actor-level supervision.
\subsection{Ablation Study}
We perform a comprehensive analysis of the different components that contribute to the effectiveness of our method. Specifically, we evaluate the impact of five individual elements:
\textbf{a)} various combinations of local and global view correspondences;
\textbf{b)} different field of view variations along the temporal and spatial dimensions;
\textbf{c)} the choice of temporal sampling strategy;
\textbf{d)} the use of spatial augmentations;
and \textbf{e)} the inference approach.

\noindent\textbf{View Correspondences}:
We propose cross-view correspondences (VC) to learn correspondences between local and global views. To investigate the effect of predicting each type of view from the other, we conduct experiments presented in Table~\ref{tbl:ablation_correspondences}. Our results show that jointly predicting $\bm{l_{t}}\to \bm{g_{t}}$ and $\bm{l_{s}}\to\bm{g_{t}}$ view correspondences leads to optimal performance. However, predicting $\bm{g_{t}}\to \bm{l_{t}}$ or $\bm{l_{s}}\to\bm{l_{t}}$ views results in reduced performance, possibly because joint prediction emphasizes learning rich context, which is absent for individual cases. We also observe a consistent performance drop for $\bm{l_{s}}\to\bm{l_{t}}$ correspondences (no overlap views), consistent with previous findings on the effectiveness of temporally closer positive views for contrastive self-supervised losses \cite{qian2020spatiotemporal, Feichtenhofer_large}.

\vspace{0.1em}
\noindent\textbf{Spatial vs. Temporal Field of View}: 
We determine the optimal combination of spatio-temporal views in Table~\ref{tbl:ablation_correspondences} by varying the field of view (crops) along both spatial and temporal dimensions (as described in Sec.\ref{subsec:cross_view_corrspondences}). To evaluate the effects of variations along these dimensions, we conduct experiments as presented in Table\ref{tbl:ablation_st}. Specifically, we compare the performance of our approach with no variation along the spatial dimension (where all frames have a fixed spatial resolution of $224\times 224$ with no spatial cropping) and with no variation along the temporal dimension (where all frames in our views are sampled from a fixed time-axis region of a video). Our findings show that temporal variations have a significant impact on NBA, while variations in the field of view along both spatial and temporal dimensions lead to the best performance (as shown in Table~\ref{tbl:ablation_st}).

\begin{figure*}[htbp!]
    \centering
    \includegraphics[width=0.70\textwidth]{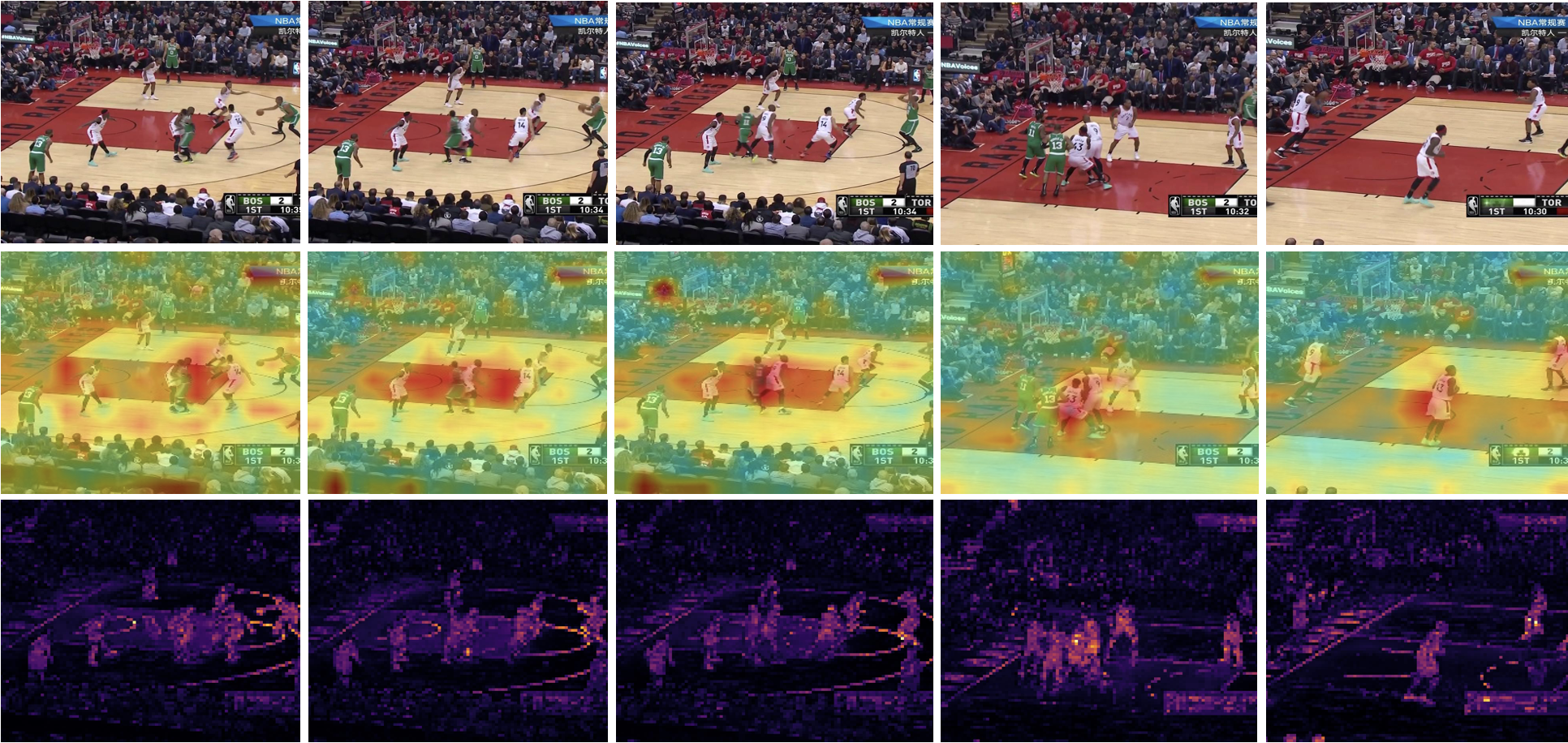}
    \vspace{-2mm}
    \caption{Visualization of the Transformer attention maps for NBA dataset. (top) Original sequence from NBA dataset~\cite{yan2020social}, (middle) Attention maps from DFWSGAR~\cite{kim2022detector} and (bottom) Attention maps from our SPARTAN model.}
    \label{fig:vis}
\end{figure*}

\begin{figure*}[!htbp]
\begin{center}
\includegraphics[width=0.9\linewidth]{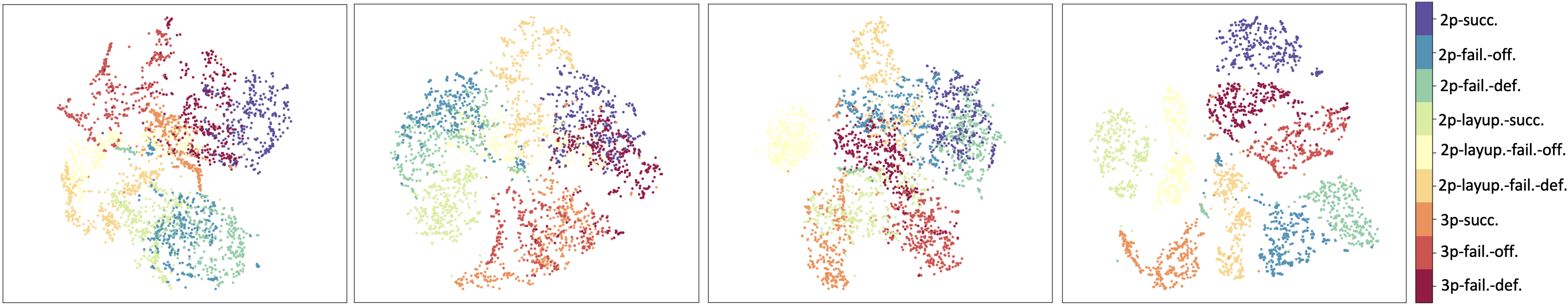}
\put(-465, 100){Base model}
\put(-340, 100){VC}
\put(-240, 100){MC + VC}
\put(-140, 100){MC + VC + SA}
\end{center}
\vspace{-1.5em}
\caption{Visualization of the $t$-SNE~\cite{van2008visualizing} plots of embedding features learned by different modules of our SPARTAN model for the NBA dataset. }
\label{fig:tsne}
\end{figure*}
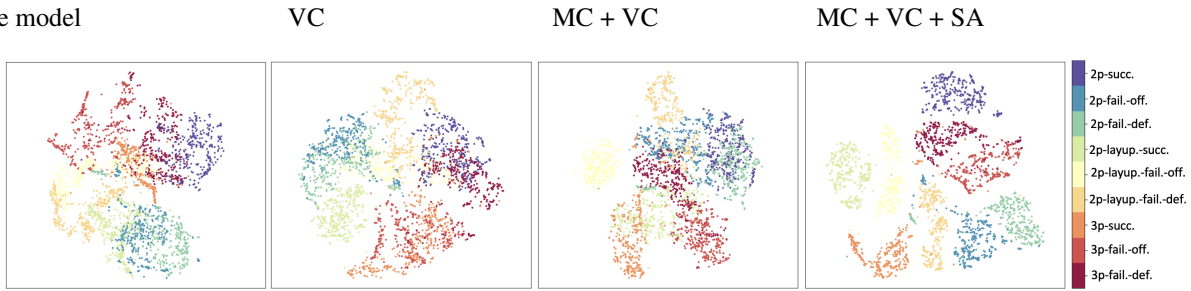

\vspace{0.1em}
\noindent\textbf{Temporal Sampling Strategy}:
Our investigation examines the possibility of replacing the temporal sampling strategy for motion correspondences (MC) proposed in our study with alternate sampling methods. To evaluate the effectiveness of MC, we replace it with an alternative approach within SPARTAN. Specifically, we test the temporal interval sampling (TIS) strategy introduced in \cite{qian2020spatiotemporal}, which has achieved state-of-the-art performance in self-supervised contrastive video settings with CNN backbones. Our experiments incorporating TIS in SPARTAN (Table \ref{tbl:ablation_temporal}) demonstrate that our proposed MC sampling strategy offers superior performance compared to TIS.
\vspace{0.1em}

\noindent\textbf{Spatial Augmentations}: 
We then investigate the impact of standard spatial augmentations (SA) on video data by experimenting with different patch sizes. Previous studies have shown that varying patch sizes can enhance the performance of CNN-based video self-supervision approaches. In our study, we evaluate the effect of patch size on our approach and present the results in Table~\ref{tbl:ablation_aug}, indicating that a patch size of 16 yields the best improvements. Based on these findings, we incorporate a patch size of 16 in our SPARTAN training process.

\vspace{0.1em}
\noindent\textbf{Inference}: 
To assess the impact of our proposed inference method (\cref{subsec:Inference}), we analyze the results presented in Table~\ref{tbl:ablation_inf}. Our findings demonstrate that our approach yields greater improvements on the NBA~\cite{yan2020social} and Volleyball~\cite{ibrahim2016hierarchical} datasets, which contain classes that can be more easily distinguished using motion information~\cite{han2020self}.
\subsection{Qualitative Results}
For the better interpretation of the results, on the NBA dataset, we visualise the attentions of the last layer of the transformer encoder in Fig.~\ref{fig:vis}. These results clearly represent the learning capability of the model towards the essential concepts, such as the position of the players along with the overall group activity classification. To demonstrate the efficiency of each module, in Fig.~\ref{fig:tsne}, we visualise the t-SNE plots. From that, we conclude that all modules combined to provide a clear seperation for each class.


\section{Conclusion}
\label{sec:conclusion}
Our work introduces SPARTAN, a self-supervised video transformer-based model. The approach involves generating multiple spatio-temporally varying views from a single video at different scales and frame rates. Two sets of correspondence learning tasks are then defined to capture the motion properties and cross-view relationships between the sampled clips. The self-supervised objective involves reconstructing one view from the other in the latent space of teacher and student networks. Moreover, our SPARTAN can model long-range spatio-temporal dependencies and perform dynamic inference within a single architecture. We evaluate SPARTAN on two group activity recognition benchmarks and find that it outperforms the current state-of-the-art models.

\noindent\textbf{Limitations:}
Our paper investigates the application of SPARTAN in the context of the RGB input modality. Currently, we do not utilize the additional supervision provided by alternate modalities in large-scale multimodal video datasets. However, in future work, we plan to explore ways in which we can modify SPARTAN to take advantage of multimodal data sources.

\small{
\noindent
\textbf{Acknowledgment} 
This work is supported by Arkansas Biosciences Institute (ABI) Grant, NSF WVAR-CRESH and NSF Data Science, Data Analytics that are Robust and Trusted (DART).
}
{\small
\bibliographystyle{ieee_fullname}

}
\end{document}